\documentclass{article} 
\usepackage{iclr2024_conference,times}

\usepackage{amsmath,amsfonts,bm}

\def\eqref#1{equation~\ref{#1}}

\def\1{\bm{1}}

\DeclareMathAlphabet{\mathsfit}{\encodingdefault}{\sfdefault}{m}{sl}
\SetMathAlphabet{\mathsfit}{bold}{\encodingdefault}{\sfdefault}{bx}{n}

\usepackage{hyperref}
\usepackage{url}
\usepackage[T1]{fontenc}
\usepackage{xcolor}
\definecolor{BrickRed}{rgb}{0.65,0.08,0.08}

\usepackage[most]{tcolorbox}  
\tcbuselibrary{skins,breakable}
\usepackage{lmodern} 
\usepackage{subcaption}

\title{Citation Parsing and Analysis with Language Models}

\author{parth sarin \\
Stanford University \\
Palo Alto, CA 94305, USA \\
\texttt{psarin@stanford.edu} \\
\And
Juan Pablo Alperin \\
Simon Fraser University \\
Vancouver, Canada \\
\texttt{juan@alperin.ca}
}

\iclrfinalcopy 
\begin{document}

\maketitle

\begin{abstract}
A key type of resource needed to address global inequalities in knowledge production and dissemination is a tool that can support journals in understanding how knowledge circulates. The absence of such a tool has resulted in comparatively less information about networks of knowledge sharing in the Global South. In turn, this gap authorizes the exclusion of researchers and scholars from the South in indexing services, reinforcing colonial arrangements that de-center and minoritize those scholars. In order to support citation network tracking on a global scale, we investigate the capacity of open-weight language models to mark up manuscript citations in an indexable format. We assembled a dataset of matched plaintext and annotated citations from preprints and published research papers. Then, we evaluated a number of open-weight language models on the annotation task. We find that, even out of the box, today's language models achieve high levels of accuracy on identifying the constituent components of each citation, outperforming state-of-the-art methods. Moreover, the smallest model we evaluated, Qwen3-0.6B, can parse all fields with high accuracy in $2^5$ passes, suggesting that post-training is likely to be effective in producing small, robust citation parsing models. Such a tool could greatly improve the fidelity of citation networks and thus meaningfully improve research indexing and discovery, as well as further metascientific research.
\end{abstract}

\section{Introduction}
Science is not an equal playing field. There is a large—though shrinking—inequality between knowledge production in the Global North and Global South. To properly understand this gap, national research capacities, and the broader power dynamics in global research contributions, it is necessary to have high quality and complete indexes of scholarly production. Accurate and complete metascientific information is also central to decision making at every level of the research system from the tenure and promotion decisions about individual faculty, to institutional resource allocation, and to national science policies. Such information is not indexed today: It has been observed for decades that research in the Global South is significantly underrepresented in supposedly global indexes, databases, and search engines for scholarly work \citep{cetto_scientific_1998,khanna_recalibrating_2022,mongeon_journal_2016}. So, for over 50 years, decisions about funding, tenure, collaboration, and governance have been made without the same type of reliable quantitative data that exists to tabulate research in the Global North. 

Fortunately, a number of projects are seeking to address global gaps in indexing and discoverability. New bibliographic databases like OpenAlex are taking a more inclusive indexing approach that has allowed them to significantly outperform existing databases in terms of coverage \citep{alperin_analysis_2024,culbert_reference_2024,jiao_how_2023}. However, despite the significantly greater coverage of works, including those from the Global South, there continues to be an enormous gap in the indexing of references \citep{alperin_analysis_2024,culbert_reference_2024}. While more complete bibliographic records are useful for understanding knowledge production, references and citations allow us to better understand the connections and circulation of knowledge, and are thus an essential aspect of research information.

This is the project to which we seek to contribute. Namely, there has been recent interest in developing new citation parsing techniques that improve on state-of-the-art methods like GROBID and Crossref search to be performant in more languages and trainable end-to-end \citep{Choi2023BuildingAA,Joshi2023AnEP}. Though they were created through very extractive and neocolonial processes \citep{gray2019ghost,hao2024ai}, modern language models can indeed operate in multiple languages and modalities and they are end-to-end trainable. Some smaller language models can also run in low-compute environments, and even on mobile devices or in the browser.

In this paper, we evaluate open-source, decoder-based language models on the task of citation parsing. Specifically, we aim to address the following research questions:
\begin{enumerate}
    \item How accurate are open-weight language models at identifying different components of a citation? How do they compare to state-of-the-art methods for reference parsing?
    \item Are small language models — which can run in the browser or on a device — promising for this task?
\end{enumerate}

\section{Methods}
\subsection{Dataset}
We began by assembling a dataset of citations in plain, formatted text, along with the same citations marked up in JATS format. We drew from two existing corpora. First, \cite{garnett_xml_2016} assembled the ``XML Markup Evaluation Corpus,'' for the Public Knowledge Project (PKP) which includes 829 submitted article manuscripts marked up in JATS format. Second, Open Research Europe (ORE) is a publishing platform with an annotated corpus of 848 articles, marked up in JATS format \citep{openresearcheurope_corpus_2025}.

We extracted the marked up version of each citation from the reference list in the JATS XML markup and matched it to a (markdown-formatted) plaintext citation extracted from the article. Specifically, we converted each article to markdown using the \texttt{markitdown} tool \citep{markitdown2025}. Then, from that markdown, we extracted plaintext citations using Llama-3.1-8B-Instruct prompted with the system prompt in Appendix~\ref{sec:citation-extraction}. We programmatically verified that each of the citations appeared in the article to prevent hallucinated citations. 

\begin{figure}[ht]
    \centering
    \begin{subfigure}[b]{0.48\linewidth}
        \centering
        \includegraphics[width=\linewidth]{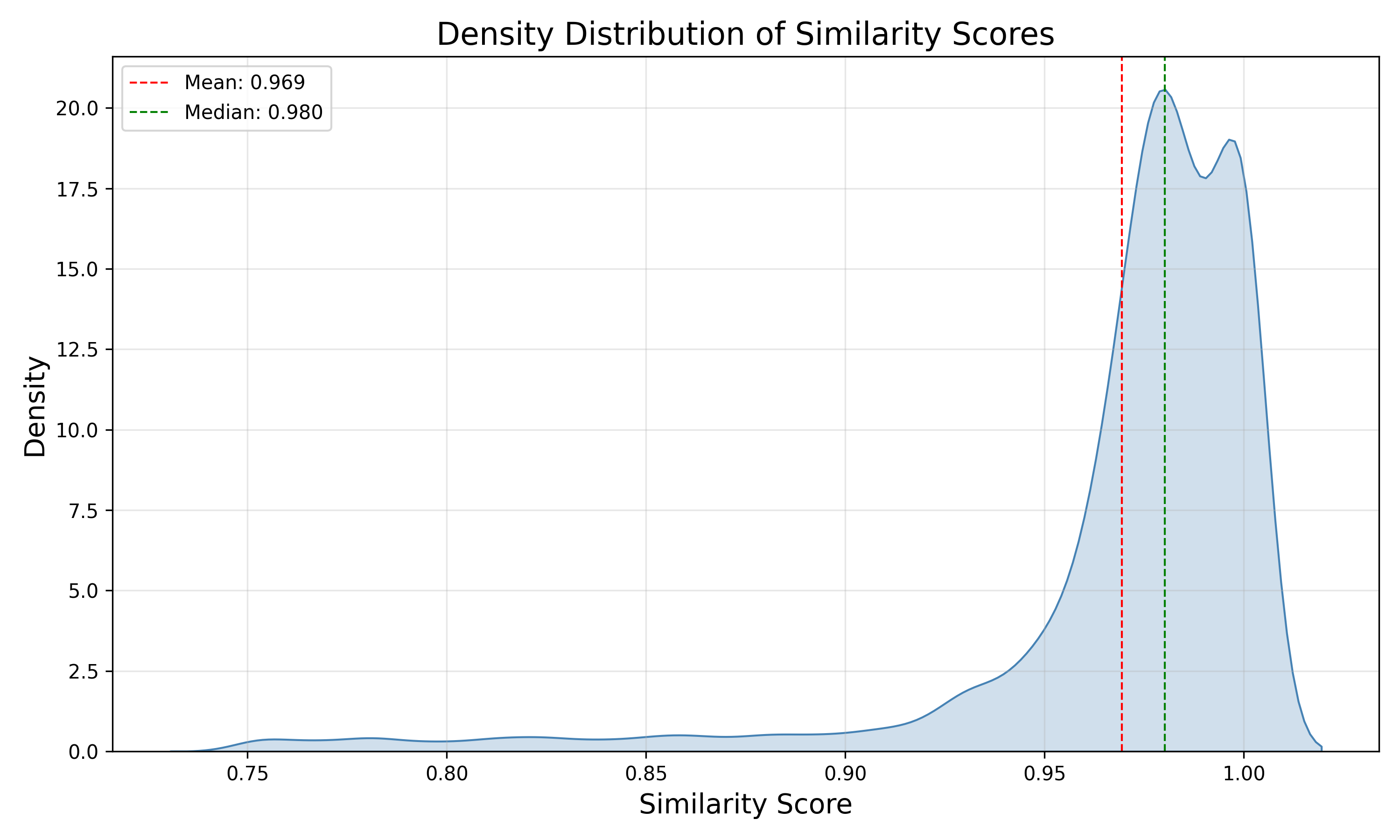}
        \caption{Garnett/PKP Corpus}
        \label{fig:pkp-similarity}
    \end{subfigure}
    \hfill
    \begin{subfigure}[b]{0.48\linewidth}
        \centering
        \includegraphics[width=\linewidth]{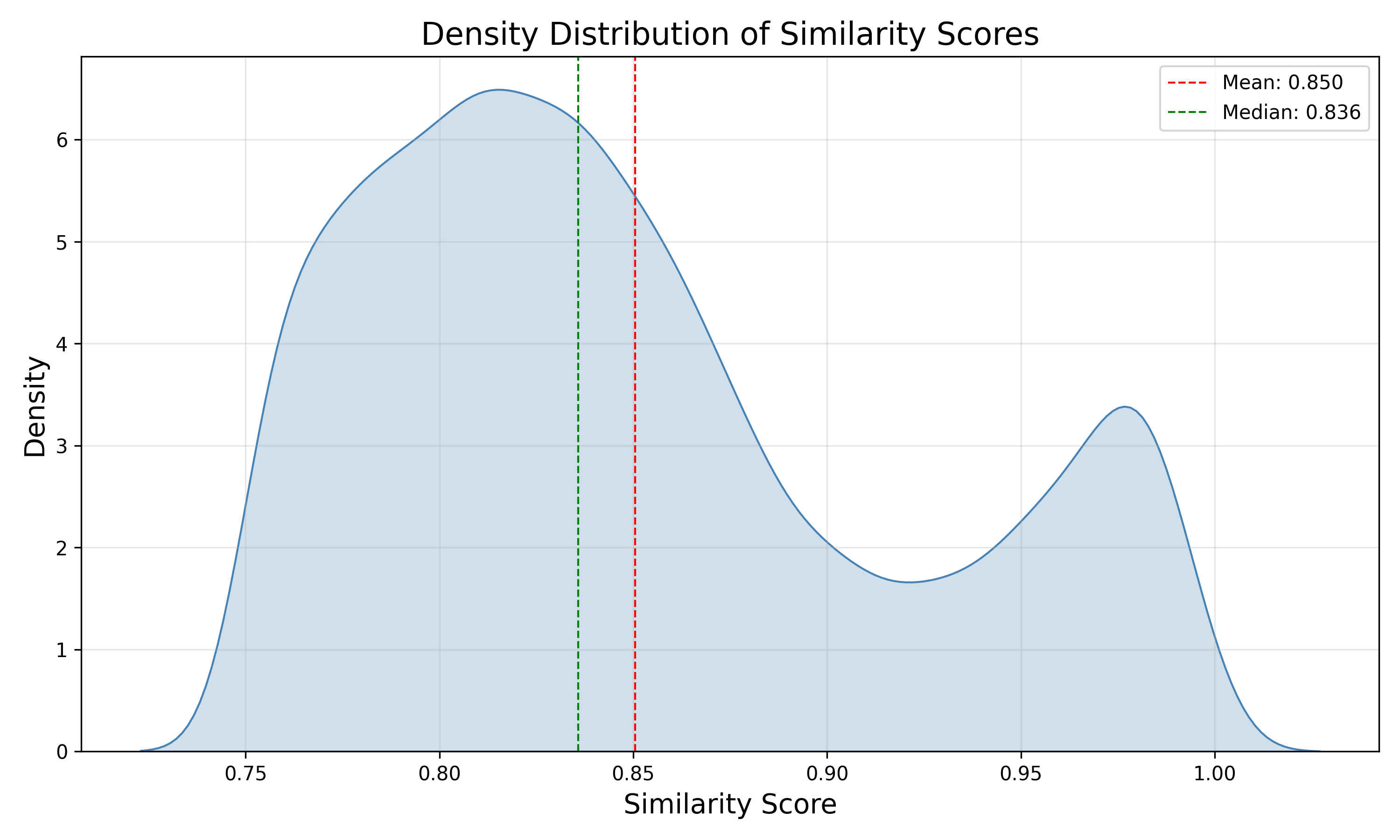}
        \caption{ORE Corpus}
        \label{fig:ore-similarity}
    \end{subfigure}
    \caption{Similarity distributions for Garnett/PKP and ORE corpora}
    \label{fig:similarity-kde}
\end{figure}

Then we matched the plaintext citations (the output of the language model) with the marked-up citations (provided in the XML files in each of the corpora). Let $p$ be the plaintext citation, $c$ be a text-only version of the marked-up citation, and $d(p, c)$ be the edit distance between the two. Take $|x|$ to represent the length (number of characters) of the string $x$. Define the \textit{similarity} between $p$ and $c$ to be \[ s(p, c) = 1 - \frac{d(p,c)}{\max\{|p|, |c|\}}. \] We matched citations in order of decreasing similarity and excluded all matches with a similarity score less than $0.75$. Figure~\ref{fig:similarity-kde} shows the distribution of similarity scores on the two corpora.

Applying this procedure, we produced 13,149 matches from the Garnett/PKP corpus and 24,934 from the ORE corpus. We sampled 1,000 citations from each for a final dataset of \textbf{2,000 matched citations} in plain text and marked up form.

We also qualitatively analyzed the matching procedure, specifically to investigate why the average similarity score is lower for the ORE corpus than the Garnett/PKP corpus. Our suspicion is that this happened because the ORE corpus is for published articles and references in that corpus can contain additional links which don't appear in the JATS markup. For example, here is one plaintext reference with a low similarity score:

\begin{tcolorbox}
Viswanathan M, Ammerman A, Eng E, et al.: Community-based participatory research: assessing the evidence. Evid Rep Technol Assess (Summ). AHRQ Publication N° 4-E022-1. Agency for Healthcare Research and Quality, Rockville, MD, 2004; (99): 1–8. \textbf{\color{BrickRed} PubMed Abstract | Free Full Text}
\end{tcolorbox}

In the published article, the highlighted section was linked to the PubMed abstract and full text of the article. The matched JATS annotation for this citation is correct but does not include those links, leading to a low score:

\begin{tcolorbox}
\begin{verbatim}
<mixed-citation publication-type="journal">
    <person-group person-group-type="author">
        <name name-style="western">
            <surname>Viswanathan</surname>
            <given-names>M</given-names>
        </name>
        <name name-style="western">
            <surname>Ammerman</surname>
            <given-names>A</given-names>
        </name>
        <name name-style="western">
            <surname>Eng</surname>
            <given-names>E</given-names>
        </name>
        <etal />
    </person-group>:
    <article-title>
        Community-based participatory research: 
        assessing the evidence.
    </article-title>
    <source>
        <italic toggle="yes">Evid Rep Technol Assess (Summ).</italic>
    </source>
    AHRQ Publication N° 4-E022-1. 
    Agency for Healthcare Research and Quality, Rockville, MD,
    <year>2004</year>; (<issue>99</issue>):
    <fpage>1</fpage>–<lpage>8</lpage>.
</mixed-citation>
\end{verbatim}
\end{tcolorbox}

\subsection{Language model evaluation}
Table~\ref{table:language-models} shows the language models we evaluated in this project. We chose small models that could potentially be deployed in low-compute contexts, given that a reference annotation service will likely need to run in the browser or on the server of a journal which may not have much compute capacity.

\begin{table}[htbp]
\centering
\caption{Model scale, maximum context length, and release date}
\begin{tabular}{lrrr}
\textbf{Model (HuggingFace repo)} &
\textbf{Params (B)} &
\textbf{Context} &
\textbf{First release} \vspace{0.1cm} \\ \hline \hline \vspace{-0.2cm} \\
deepseek-ai/DeepSeek-R1-Distill-Qwen-14B & 14 & 32,768 & Feb 2025\\
deepseek-ai/DeepSeek-R1-Distill-Qwen-7B  & 7  & 32,768 & Feb 2025\\
Qwen/Qwen2.5-7B-Instruct                 & 7.6 & 131,072 & Apr 2025\\
Qwen/Qwen2.5-3B-Instruct                 & 3.1 & 32,768  & Apr 2025\\
Qwen/Qwen3-4B                            & 4.0 & 32,768 & May 2025\\
Qwen/Qwen3-1.7B                          & 1.7 & 32,768             & May 2025\\
Qwen/Qwen3-0.6B                          & 0.6 & 32,768             & Apr 2025\\
microsoft/phi-4                          & 14  & 16,000              & Dec 2024\\
microsoft/Phi4-mini-instruct             & 3.8 & 128,000             & Feb 2025\\
meta-llama/Llama-3.1-8B-Instruct         & 8   & 128,000             & Jul 2024\\
meta-llama/Llama-3.2-3B-Instruct         & 3   & 128,000             & Sep 2024\\
\hline
\end{tabular}
\label{table:language-models}
\end{table}

Each model was prompted with two examples of marked-up citations (from the larger dataset, not the sampled 2,000) and instructed to produce the XML citation for a particular citation. We prompted the model twice, with and without a chain of thought. For the reasoning models, we explicitly foreclosed its reasoning by appending \verb|<think></think>| to the end of the prompt. When we sampled a chain of thought, we sampled at temperature $0.6$, with top $p$ set to $0.95$ and top $k$ set to $20$. For the non-CoT prompts we sampled at temperature $0.7$ with a top $p$ value of $0.8$ and top $k$ of $20$. In the results, we report the \verb|pass@1| accuracies for each model.

We evaluated the language model's accuracy on a number of citation sub-elements: \verb|article-title|, \verb|issue|, \verb|volume|, \verb|source|, \verb|coverage|, \verb|year|, \verb|fpage|, and \verb|surname| (the first author's surname). For each field except \verb|article-title| and \verb|source|, we required that the prediction exactly match the label. We allowed for an edit distance of $10$ for the article title and $5$ for the source. We also recorded the coverage, or the percent of citations for which the language model produced valid XML. If the language model did not produce valid XML, its label was marked incorrect for all sub-elements.

We also sampled $2^6 = 64$ completions for each citation, with a reasoning trace, from Qwen3-0.6B to determine whether this task is in distribution for a small model.

Running these experiments (\verb|pass@1| accuracies for eleven models and \verb|pass@64| accuracy for one model) took six hours on an H200 GPU. Based on our estimates, including the CO$_2$ emissions of the GPU and the datacenter, these experiments required at most 2.46 kg CO$_2$, or roughly the emissions incurred by driving the average U.S. gasoline-powered passenger vehicle 6.3 miles. See Appendix~\ref{sec:energy-consumption} for a more detailed calculation.

\section{Results}
\subsection{RQ1: Model accuracies}
\begin{figure}[ht]
    \centering
    \includegraphics[width=0.9\linewidth]{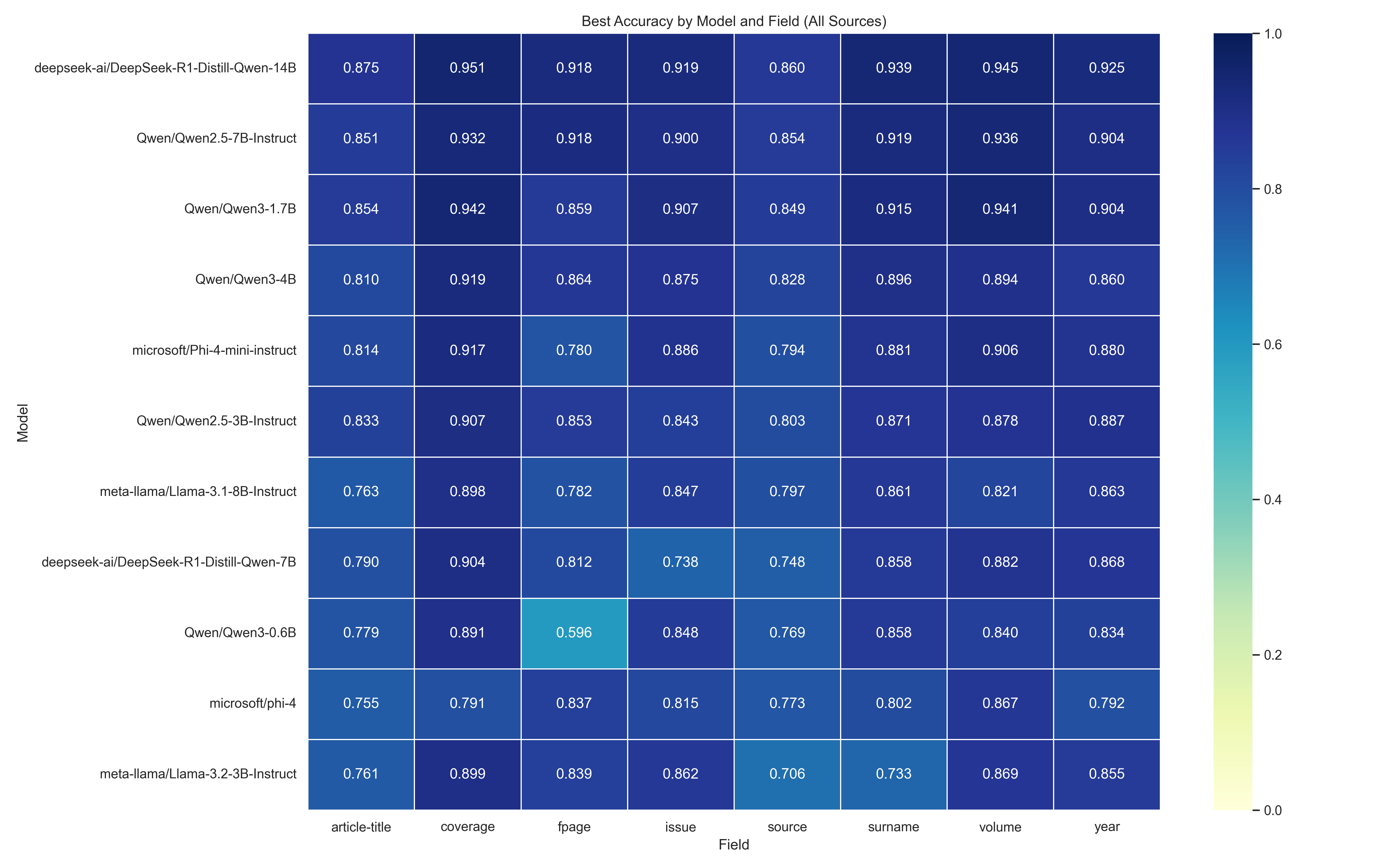}
    \caption{\texttt{pass@1} model accuracies, max between CoT and non-CoT prompts}
    \label{fig:model-accuracies}
\end{figure}

Figure~\ref{fig:model-accuracies} shows the accuracies of these models on the 2,000 citations. We report the highest \verb|pass@1| accuracy for each field across the two prompts (with and without a chain of thought).

We also conducted a quick evaluation of Crossref search for these references. Crossref had coverage of $0.968$ with an article title accuracy of $0.439$. The API also provides a ``confidence'' score. Filtering only to high confidence items (a score of larger than $50$), this approach has a coverage of $0.632$ with an article title accuracy of $0.633$. Both of these accuracies are lower than the worse-performing language models.

We did a similar evaluation of GROBID using the \verb|/api/processCitation| endpoint. It had a coverage of $0.989$ with an article title accuracy of $0.6674$ and a surname accuracy of $0.8516$. The worst performing language model does better by $8.76$ percentage points for article title accuracy and most language models (except for the worst two) also have a higher accuracy for surname identification.

\subsection{RQ2: Small language models}\label{sec:small-lm-distribution}
\cite{yue2025doesreinforcementlearningreally} demonstrated that the capacity of a language model to learn a task using reinforcement learning with verifiable reward (RLVR) is strongly connected to the base model's \verb|pass@k| performance on the task. This provides one technique to determine whether high-fidelity citation parsing is within the distribution of a small model.

\begin{figure}[ht]
    \centering
    \includegraphics[width=0.9\linewidth]{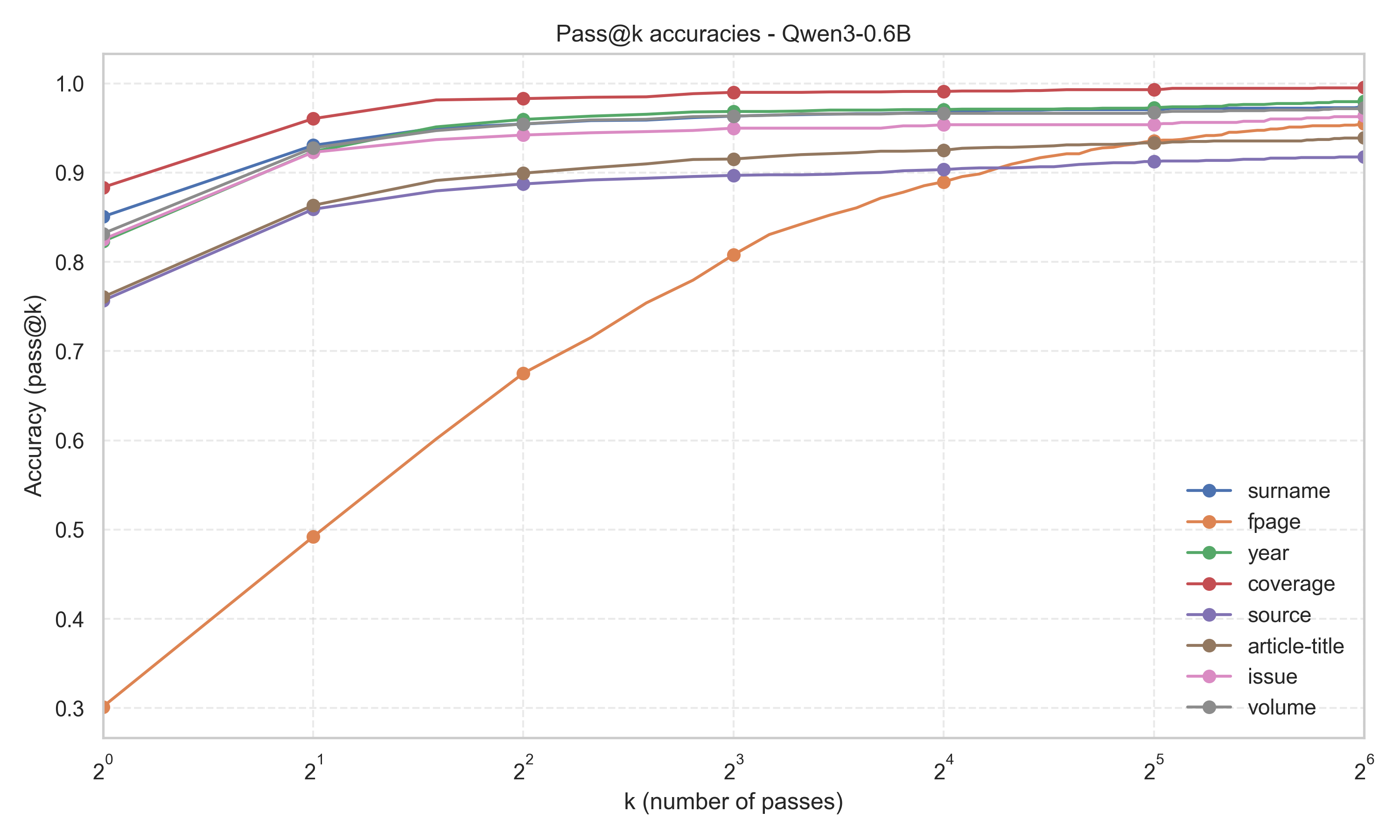}
    \caption{\texttt{pass@64} accuracy for Qwen3-0.6B on citation parsing}
    \label{fig:qwen3-0.6b-pass@k}
\end{figure}

Figure~\ref{fig:qwen3-0.6b-pass@k} shows whether the correct answer for each sub-element appears within $64$ samples, with reasoning, from Qwen3-0.6B. It shows that this task (over 90\% accuracy for all sub-elements) is within distribution for the model and that the standard RLVR setup is likely to improve sampling efficiency.

\section{Discussion}
This short report has demonstrated the capacity of language models to achieve high accuracy at marking up the different components of a plaintext citation. Specifically, the models evaluated in this study generally achieve higher field accuracy than state-of-the-art methods like GROBID and Crossref search. It also seems within the capacity of even the smallest model we tested (Qwen3-0.6B) to achieve high \verb|pass@1| accuracy with post-training through reinforcement learning. In this section, we conclude by discussing a few further directions for this research.

First, deploying these models will likely require specialized compute infrastructure, further model optimization, or both. Although, in this work, we tested non-quantized models on the latest GPUs, additional research is needed to evaluate accuracy and feasibility in the low-resource context that is typically available for academic journals.

Also, although the test in Section~\ref{sec:small-lm-distribution} shows that RLVR is likely to be effective for Qwen3-0.6B, sampled with reasoning, there are a variety of other training methods that could be used which are more able to shift the underlying distribution of a language model. In particular, distillation and supervised fine-tuning may be effective in learning a high-accuracy annotation system, especially because this problem has a lot of available training data.

Similarly, we did not explore any test-time techniques that could have improved accuracy or coverage. For instance, one can define a grammar and use constrained decoding to guarantee that the language model produces valid JATS XML, boosting coverage to 100\% \citep{Scholak2021PICARDPIA,poesia2022synchromeshreliablecodegeneration}. It has also been demonstrated that Best-of-N sampling, process supervision with reward models, and self-feedback can improve accuracy at inference time \citep{snell2024scalingllmtesttimecompute,madaan2023selfrefineiterativerefinementselffeedback}. Most of these techniques either require more compute or are not well optimized for existing inference frameworks so may not make sense for deployment.

One other limitation of our work is that it relies on an initial extraction phase where we used Llama-3.1-8B-Instruct to identify the plaintext list of citations. We did not test the accuracy of this step, but we did validate that all of the matched citations have high similarity. Another direction for future research is to evaluate extraction, which will be important for offline metadata enrichment.

Finally, it is worth returning to the significance of a tool which can efficiently and accurately parse the elements of a citation. Not only could this tool be deployed as part of journal software, but it could be applied offline to enrich the metadata for existing research publications. This new infrastructure could shed more light on how knowledge circulates in languages and contexts that have systematically excluded from the dominant research communities of the Global North.

\section{Funding}
This research was supported by free compute credits provided by \href{https://modal.com/}{Modal} and by the Social Sciences and Humanities Research Council of Canada (SSHRC) through Grant \#1007-2023-0001. The funder had no role in the study design, data collection and analysis, decision to publish, or preparation of the manuscript. 

The authors declare no conflicts of interest.

\newpage
\bibliography{iclr2024_conference}
\bibliographystyle{iclr2024_conference}

\newpage
\appendix
\section{Prompts}
\subsection{Citation extraction}\label{sec:citation-extraction}
\begin{tcolorbox}[title=System prompt for citation extraction]
You are an expert annotator that specializes in reading academic articles and isolating their bibliography. You will see the text of an academic article and you should write out a list of the references from the article bibliography, one on each line. Copy the references verbatim. Respond only with the list of references, no other text. Do not include bullet points or numbered lists, but edit the formatting so each citation fits on exactly one line. Do not include parenthetical citations. Just focus on the bibliography section.
\end{tcolorbox}

\subsection{Few-shot annotation}\label{sec:few-shot-prompts}
\begin{tcolorbox}[
  title=No chain-of-thought system prompt,
  enhanced,
  breakable,
  halign=left,
  parbox=false,                       
  before upper={\setlength{\parskip}{6pt}}, 
  skin first=enhanced,
  skin middle=enhanced,
  skin last=enhanced
]
You are a helpful assistant with an expertise in annotating bibliographic references in JATS XML format.

You will be given a plaintext citation and you should respond with the annotated reference. ONLY respond with the annotation and nothing else.

\# Examples

\#\#\# Citation

Stone NJ, Robinson JG, Lichtenstein AH, et al. 2013 ACC/AHA Guideline on the Treatment of Blood Cholesterol to Reduce Atherosclerotic Cardiovascular Risk in Adults: A Report of the American College of Cardiology/American Heart Association Task Force on Practice Guidelines. Circulation 2014;129(25 Suppl 2):S1-S45. doi:10.1161/01.cir.0000437738.63853.7a.

\#\#\# Annotation

<mixed-citation publication-type="journal"><person-group person-group-type="author"><string-name><surname>Stone</surname> <given-names>NJ</given-names></string-name>, <string-name><surname>Robinson</surname> <given-names>JG</given-names></string-name>, <string-name><surname>Lichtenstein</surname> <given-names>AH</given-names></string-name>, <etal>et al</etal></person-group>. <article-title>2013 ACC/AHA Guideline on the Treatment of Blood Cholesterol to Reduce Atherosclerotic Cardiovascular Risk in Adults: A Report of the American College of Cardiology/American Heart Association Task Force on Practice Guidelines</article-title>. <source><italic>Circulation</italic></source> <year>2014</year>;<volume>129</volume>(<issue>25</issue> <supplement>Suppl 2</supplement>):<fpage>S1</fpage>-<lpage>S45</lpage>. <comment>doi</comment>:<pub-id pub-id-type="doi">10.1161/01.cir.0000437738.63853.7a</pub-id>.</mixed-citation>

\#\#\# Citation

Sagel Z, Tutluer Mİ, Peskircioglu H, et al.: Determination of Effect of Chemical Mutagen EMS on TAEK A-3 and TAEK C-10 Mutant Soybean Varieties in M1 Generation. Ekin Journal of Crop Breeding and Genetics. 2017; 3(1): 19–24. Reference Source

\#\#\# Annotation

<mixed-citation publication-type="journal"><person-group person-group-type="author"><name name-style="western"><surname>Sagel</surname> <given-names>Z</given-names></name>, <name name-style="western"><surname>Tutluer</surname> <given-names>Mİ</given-names></name>, <name name-style="western"><surname>Peskircioglu</surname>, <given-names>H</given-names></name>, <etal /></person-group>:<article-title>Determination of Effect of Chemical Mutagen EMS on TAEK A-3 and TAEK C-10 Mutant Soybean Varieties in M<sub>1</sub> Generation.</article-title>. <source><italic toggle="yes">Ekin Journal of Crop Breeding and Genetics</italic></source>. <year>2017</year>;<volume>3</volume>(<issue>1</issue>): <fpage>19</fpage>–<lpage>24</lpage>. <ext-link>Reference Source</ext-link></mixed-citation>
\end{tcolorbox}

\begin{tcolorbox}[
title=Chain-of-thought system prompt,
  enhanced,
  breakable,
  halign=left,
  parbox=false,                       
  before upper={\setlength{\parskip}{6pt}}, 
  skin first=enhanced,
  skin middle=enhanced,
  skin last=enhanced
]
You are a helpful assistant with an expertise in annotating bibliographic references in JATS XML format.

You will be given a plaintext citation and you should respond with the annotated reference. Before providing the final XML annotation, give your step-by-step thinking. After your explanation, provide the final XML annotation.

\# Examples

\#\#\# Citation

Stone NJ, Robinson JG, Lichtenstein AH, et al. 2013 ACC/AHA Guideline on the Treatment of Blood Cholesterol to Reduce Atherosclerotic Cardiovascular Risk in Adults: A Report of the American College of Cardiology/American Heart Association Task Force on Practice Guidelines. Circulation 2014;129(25 Suppl 2):S1-S45. doi:10.1161/01.cir.0000437738.63853.7a.

\#\#\# Annotation

<mixed-citation publication-type="journal"><person-group person-group-type="author"><string-name><surname>Stone</surname> <given-names>NJ</given-names></string-name>, <string-name><surname>Robinson</surname> <given-names>JG</given-names></string-name>, <string-name><surname>Lichtenstein</surname> <given-names>AH</given-names></string-name>, <etal>et al</etal></person-group>. <article-title>2013 ACC/AHA Guideline on the Treatment of Blood Cholesterol to Reduce Atherosclerotic Cardiovascular Risk in Adults: A Report of the American College of Cardiology/American Heart Association Task Force on Practice Guidelines</article-title>. <source><italic>Circulation</italic></source> <year>2014</year>;<volume>129</volume>(<issue>25</issue> <supplement>Suppl 2</supplement>):<fpage>S1</fpage>-<lpage>S45</lpage>. <comment>doi</comment>:<pub-id pub-id-type="doi">10.1161/01.cir.0000437738.63853.7a</pub-id>.</mixed-citation>

\#\#\# Citation

Sagel Z, Tutluer Mİ, Peskircioglu H, et al.: Determination of Effect of Chemical Mutagen EMS on TAEK A-3 and TAEK C-10 Mutant Soybean Varieties in M1 Generation. Ekin Journal of Crop Breeding and Genetics. 2017; 3(1): 19–24. Reference Source

\#\#\# Annotation

<mixed-citation publication-type="journal"><person-group person-group-type="author"><name name-style="western"><surname>Sagel</surname> <given-names>Z</given-names></name>, <name name-style="western"><surname>Tutluer</surname> <given-names>Mİ</given-names></name>, <name name-style="western"><surname>Peskircioglu</surname>, <given-names>H</given-names></name>, <etal /></person-group>:<article-title>Determination of Effect of Chemical Mutagen EMS on TAEK A-3 and TAEK C-10 Mutant Soybean Varieties in M<sub>1</sub> Generation.</article-title>. <source><italic toggle="yes">Ekin Journal of Crop Breeding and Genetics</italic></source>. <year>2017</year>;<volume>3</volume>(<issue>1</issue>): <fpage>19</fpage>–<lpage>24</lpage>. <ext-link>Reference Source</ext-link></mixed-citation>
\end{tcolorbox}

\newpage
\section{Energy consumption}\label{sec:energy-consumption}
NVIDIA lists the max thermal-design power (TDP) of the H200 chip to be up to \textbf{700W} \citep{nvidia_h200_2024}. Assuming that it was run at its full TDP for six hours, it would have used $0.7 \text{ kW} \times 6\text{ h} = 4.2 \text{ kWh}$ during its use. In 2024, the average power use effectiveness (PUE) for a U.S.-based center was $1.56$ \citep{uptime_survey_2024}. Assuming this PUE, we can estimate that our experiments actually required $4.2 \text{ kWh} \times 1.56 \approx \mathbf{6.6 \textbf{ kWh}}$.

The average carbon intensity for the U.S. power grid is estimated to be ${823.1}$ lb CO$_2$ per mWh \citep{epa_ghg_equivalencies_2024}. This is equivalent to $373.35$ kg CO$_2$ per mWH, or \textbf{$\mathbf{0.373}$ kg CO$\mathbf{_2}$ per kWh}.

Thus, we estimate the overall energy consumption of these experiments to be $6.6 \text{ kWh} \times 0.373\text{ kg CO$_2$/kWh} \approx \mathbf{2.46\textbf{ kg CO$_2$}}$.

According to the EPA, the average U.S. gasoline-powered passenger vehicle (including cars and trucks) emits $3.93 \times 10^{-4}$ metric tons CO$_2$ per mile, or $0.393$ kg CO$_2$ per mile \citep{epa_ghg_equivalencies_2024}. This means that the total emissions for our experiments are roughly equivalent to driving this vehicle \[ \frac{2.46 \text{ kg CO$_2$}}{0.393\text{ kg CO$_2$ per mile}} \approx \mathbf{6.26 \textbf{ miles}}. \]
\end{document}